\documentclass[letterpaper, 10 pt, conference]{ieeeconf}

\usepackage{amsmath, amssymb}
\usepackage{amsfonts}
\usepackage{microtype}
\usepackage{hyperref}
\newcommand\norm[1]{\left\lVert#1\right\rVert}
\usepackage{txfonts}
\usepackage{graphicx,bm,float}
\usepackage{subcaption}
\IEEEoverridecommandlockouts                              

\title{\LARGE \bf
\textit{Footstep Planning with Encoded Linear Temporal Logic Specifications \\}
}

\author{Vikram Ramanathan
}

\begin{document}

\maketitle
\thispagestyle{empty}
\pagestyle{empty}

\begin{abstract}
This article presents an approach to encode Linear Temporal Logic (LTL) Specifications into a Mixed Integer Quadratically Constrained Quadratic Program (MIQCQP) footstep planner. We propose that the integration of LTL specifications into the planner not only facilitates safe and desirable locomotion between obstacle-free regions, but also provides a rich language for high-level reasoning in contact planning. Simulations of the footstep planner in a 2D environment satisfying encoded LTL specifications demonstrate the results of this research.

\end{abstract}

\section{Introduction}
Humanoid locomotion is accomplished by changing its footholds. In order to move feasibly and efficiently, a footstep planner must be able to find sequences of foot positions and orientations that will realize the robot's locomotion objectives. One established solution for the footstep planning problem is to perform some discrete sampling-based search algorithm over reachable foot positions and orientations. This approach first determines candidate footsteps through one of several methods, including:

\begin{itemize}
    \item intersecting valid sample limb configurations with the environment to determine which limb configurations are close to contact and then projecting these configurations onto contact using inverse kinematics \cite{tonneau2018efficient}
    \item assuming general bounds on the displacement of each foothold based on the robot structure and kinematic reachability \cite{hornung2012anytime} \cite{karkowski2016real} \cite{kuffner2005motion} \cite{baudouin2011real}
    \item defining an action-effect mapping \cite{asimoplanner2005}
\end{itemize}

Then, a search algorithm, such as A* \cite{michel2005vision} \cite{asimoplanner2005} \cite{lin2019efficient} \cite{kuffner2005motion} or Rapidly-exploring Random Trees (RRT) \cite{baudouin2011real}  \cite{xia2010global} is used to find a sequence of footsteps. However, the performance of these search algorithms largely depends on the quality of the heuristic function that guides the search as well as the granularity of the discretization. As an alternative, Deits and Tedrake  \cite{deits2014footstep} proposed a footstep planner formulated as a mixed integer optimization program with continuous decision variables for footstep position and orientation, and integer decision variables to eliminate non-convex constraints. After employing the IRIS algorithm \cite{Deits2015computing} to compute large obstacle-free convex regions within the provided environment, their solution solves a Mixed Integer Quadratically Constrained Quadratic Program (MIQCQP) with a clever discretization of \textit{cos} and \textit{sin} in its reachability constraint. This approach solves for globally optimal footstep plans at realtime rates while satisfying reachability constraints and avoiding obstacles.

\section{Problem Description}

In this article, we consider the problem of encoding temporal logic specifications into a MIQCQP footstep planner to improve and extend its functionality. Continuous footstep planners like the Mixed Integer Quadratically Constrained Quadratic Program (MIQCQP) footstep planner proposed by Deits and Tedrake \cite{deits2014footstep} adopt an optimization framework, thereby avoiding the use of search heuristics as well as the sampling of discretized footstep configurations. The problem of integrating temporal logic specifications into a sampling-based search algorithm such as Probabilistic Road Maps (PRMs) and Randomly-Exploring Random Trees (RRTs) has been studied \cite{vasile2013sampling} \cite{karaman2009sampling}. However, we have adopted the continuous footstep planner to take advantage of its ability to solve for optimal solutions as well as the ease with which its optimization framework can be expanded upon to include simplified robot dynamics constraints and compute for reaction forces.

\section{Solution}
As mentioned previously, this work builds upon Deits' \cite{deits2014footstep} footstep planner. This planner solves an MIQCQP to determine x, y and $\theta$ for N footsteps. We first will cover the formulation of this footstep planner before discussing the encoding of LTL specifications. 
\subsection{MIQCQP Footstep Planner}
The footstep planner seeks to minimize a quadratic cost function (in terms of the length of each stride and the distance between the terminal footstep and the goal footstep position and orientation) subject to obstacle avoidance and reachability constraints. Obstacle avoidance is facilitated by first decomposing the non-convex obstacle-free configuration space into a set of convex ``safe" regions \cite{Deits2015computing}. These convex regions are then defined as polygons by the halfspace respresentation and added as convex constraints to the program. A binary variable is assigned to each possible footstep and region pair. If such a binary variable is 1, the corresponding footstep is assigned to the region. More precisely, the matrix, $H \in \{0, 1\}^{R\times N}$, where $R$ is the number of regions and $N$ is the number of footsteps, is constructed such that if $H_{r,j}$ = 1, then the $j^{th}$ footstep, $j = \{1,...,N\}$, is assigned to the region $r$. This implication is defined as a mixed-integer linear constraint using the big-M formulation \cite{vielma2015mixed},
\begin{equation}
    -M (1 - H_{r,j}) + A_r f_j \leq b_r
\end{equation}

where $$ f_j= \begin{bmatrix} x_j \\ y_j \\ \theta_j  \end{bmatrix}$$, $M$ is a sufficiently large positive number, $A_r$ defines the normal vectors corresponding to the supporting halfplanes of the polygonal obstacle-free regions, and $b_r$ defines the offsets between the halfplanes and the origin. However, since the footstep must only belong to one region instead of the intersection of all these regions, $\sum_{r=1}^{R} H_{r,j} = 1$ must be enforced. 

The footstep reachability constraint is defined as the intersection of two circular regions offset from the previous foothold (as shown in Eq. \ref{eqn:footstep_reach} below). These offsets are mirrored for left and right feet.

\begin{equation}
     \norm{ \begin{bmatrix}x_j \\ y_j \end{bmatrix} - \bigg( \begin{bmatrix}x_{j-1} \\ y_{j-1} \end{bmatrix} + \begin{bmatrix} cos(\theta_j) & -sin(\theta_j) \\ sin(\theta_j) & cos(\theta_j) \end{bmatrix} p_i \bigg) } \leq r_i 
     \label{eqn:footstep_reach}
\end{equation}

where $i \in \{1,2\}$, $j \in \{2,...,N\}$, $p_i$ are the centers of the circles, and $r_i$ are the radii of the circles. By tuning $p_i$ and $r_i$, we can obtain a conservative approximation of a biped's footstep reachability. In order to eliminate the nonlinearity, $cos$ and $sin$ are approximated by linear piece-wise functions along with binary variables to decide which linear segment of the function to use based on $\theta$'s value. These binary variables are defined as binary matrices, $S, C \in \{0, 1\}^{L \times N}$, where L is the number of piece-wise segments and N is the number of footsteps. Since $\theta$ cannot belong to multiple segments, $\sum_{l=1}^{L} S_{l,j}= \sum_{l=1}^{L} C_{l,j} = 1$ must be satisfied. Our piece-wise approximation of $cos$ and $sin$ defines 5 linear segments:

\begin{equation}
    sin(\theta) = \begin{cases} -\theta - \pi \hskip 1.5em -\pi \leq \theta < 1-\pi \\
    -1 \hskip 3.8em 1 \leq \theta < -1\\
    \theta \hskip 4.5em -1 \leq \theta < 1 \\
    1 \hskip 4.5em 1 \leq \theta < \pi - 1\\
    -\theta + \pi \hskip 2em \pi-1 \leq \theta < \pi \\
    \end{cases}
\end{equation}

\begin{equation}
    cos(\theta) = \begin{cases} -1 \hskip 4.3em -\pi \leq \theta < -\pi/2-1 \\
    \theta + \pi/2 \hskip 2.3em -\pi/2-1 \leq \theta < 1 - \pi/2\\
    1 \hskip 5.3em 1-\pi/2 \leq \theta < \pi/2 - 1 \\
    -\theta + \pi/2 \hskip 2em \pi/2 -1 \leq \theta < \pi/2 + 1\\
    -1 \hskip 4.75em \pi/2 + 1 \leq \theta < \pi \\
    \end{cases}
\end{equation}
\begin{figure*}[t!]
    \centering
    \begin{subfigure}[t]{0.5\textwidth}
        \centering
        \includegraphics[height=2.5in]{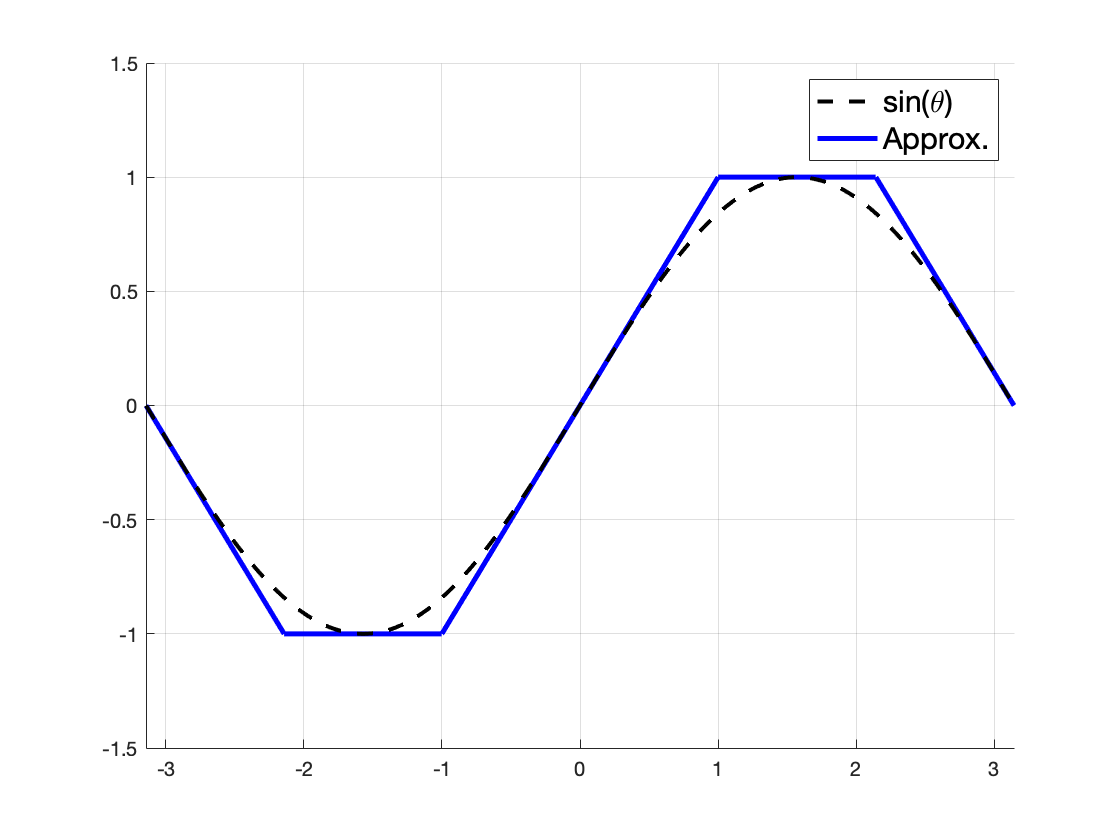}
        \caption{$sin(\theta)$}
    \end{subfigure}%
    ~ 
    \begin{subfigure}[t]{0.5\textwidth}
        \centering
        \includegraphics[height=2.5in]{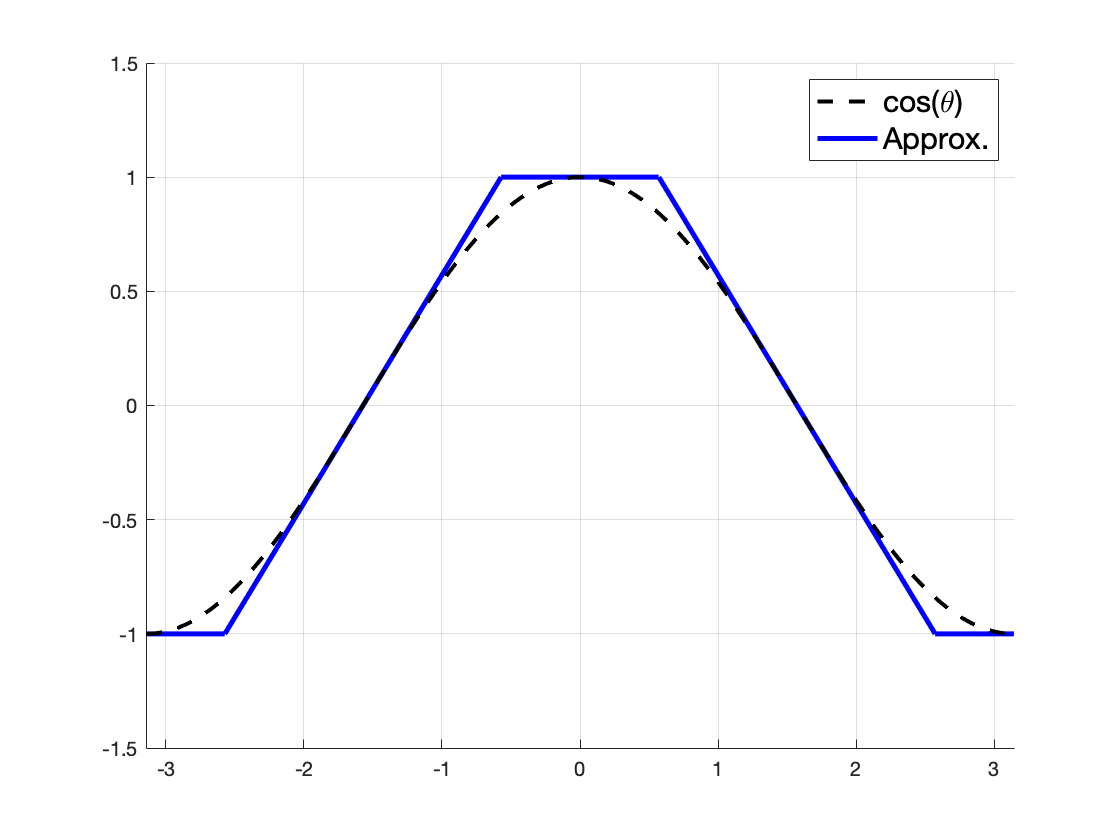}
        \caption{$cos(\theta)$}
    \end{subfigure}
    \caption{Piece-wise Approximations of $cos(\theta)$ and $sin(\theta)$}
    \label{fig:approx}
\end{figure*}

The graphical representation of this approximation is shown in Fig. \ref{fig:approx}. We also bound the change in footstep orientation in every step to $\pi/8$ rad. 

Putting the quadratic cost function and all these constraints together, we get the following MIQCQP:

\begin{equation}
    \begin{aligned}
        & \underset{f_1,...,f_j,S,C,H}{\text{minimize}}
        && (f_N - g)^TQ(f_N-g) + \sum_{j=1}^{N-1}(f_{j+1} - f_j)^TR(f_{j+1}-f_j) \\
        & \text{subject to}
        && H_{r,j} \implies A_rf_j \leq b_r \\
        &&& S_{l,j} \implies \begin{cases} 
                          \phi_l \leq \theta_j \leq \phi_{l+1} \\
                          s_j = g_l \theta_j + h_l \\
                       \end{cases} \\
        &&& C_{l,j} \implies \begin{cases} 
                          \phi_l \leq \theta_j \leq \phi_{l+1} \\
                          c_j = g_l \theta_j + h_l \\
                       \end{cases} \\
        &&& \norm{ \begin{bmatrix}x_j \\ y_j \end{bmatrix} - \bigg( \begin{bmatrix}x_{j-1} \\ y_{j-1} \end{bmatrix} + \begin{bmatrix} c_j & -s_j \\ s_j & c_j \end{bmatrix} p_i \bigg) } \leq r_i \\
        &&& \sum_{r=1}^{R}H_{r,j} = \sum_{l=1}^{L}S_{l,j} = \sum_{l=1}^{L}C_{l,j} = 1 \\
        &&& \theta_{j} - \theta_{j-1} \leq \pi/8
    \end{aligned}
\end{equation}

where $r = 1...R$, , $j = \{1,...,N\}$, and $l = 1,...,L$.

\subsection{Linear Temporal Logic}
\subsubsection{Preliminaries}
In this work, the underlying time domain is discrete since each ``execution" or footstep corresponds to the advancement of a single time-unit. We consider Linear Temporal Logic (LTL) to describe desired planning behavior. LTL formulae are defined over a set of atomic propositions, $AP$, according to the following grammar \cite{baier2008principlesmodel}:

$$ \phi ::= \text{true } | \,a \, | \, \phi_1 \wedge \phi_2 \, | \, \lnot \phi \, | \, \bigcirc \phi \, | \, \phi_1 U \phi_2  $$

where $a \in AP$, $\bigcirc$ and $U$ are the temporal operators for ``next" (signifies that the proposition is true in the next step) and ``until" (signifies that the proposition preceding the operator is true until the future moment when the proposition succeeding the operator is true), and $\lnot$ and $\wedge$ are the negation and conjuction Boolean operators respectively. The other Boolean operators for disjunction, $\vee$, implication $\implies$, and equivalence $\iff$ can be derived from the previously defined grammar:
\begin{equation}
    \phi_1 \vee \phi_2 = \lnot(\lnot \phi_1 \wedge \lnot \phi_2)
    \label{eqn:disjunc_equiv}
\end{equation}
\begin{equation}
    \phi_1 \implies \phi_2 = \lnot \phi_1 \vee \phi_2
\end{equation}
\begin{equation}
    \phi_1 \iff \phi_2 = (\phi_ 1 \implies \phi_2) \wedge (\phi_2 \implies \phi_1)
\end{equation}

In addition, the temporal operators, $\lozenge$ (eventually) and $\square$ (always) are defined by:
\begin{equation}
    \lozenge \phi = true \, U \, \phi
\end{equation}
\begin{equation}
    \square \phi = \lnot\lozenge\lnot\phi
\end{equation}

Let $\sigma$ be an infinite sequence of states, $\sigma = A_0, A_1, ... (2^{AP})^\omega$. Let $\sigma_i$ denote the word $\sigma$ from position $i$. The semantics of an LTL formula is defined as a language $Words(\phi)$ that contains all infinite words over the alphabet $2^{AP}$ that satisfy $\phi$. Every LTL formula is associated with a single linear time property. The satisfaction relations are defined inductively by:
\begin{equation}
    \sigma_i \models a \; \text{iff} \; a \in A_0
    \label{eqn:sat1}
\end{equation}
\begin{equation}
    \sigma_i \models \phi_1 \wedge \phi_2 \; \text{iff} \; \sigma_i \models \phi_1 \; \text{and} \; \sigma_i \models \phi_2
\end{equation}
\begin{equation}
    \sigma_i \models \lnot \phi \; \text{iff} \; \sigma_i \nvDash \phi
\end{equation}
\begin{equation}
    \sigma_i \models \bigcirc \phi \; \text{iff} \; \sigma_{i+1} \models \phi
\end{equation}
\begin{multline}
    \sigma_i \models \phi_1 \; U \; \phi_2 \; \text{iff} \; \exists j \geq i \; \text{s.t.} \; \sigma_j \models \phi_2 \; \text{and} \; \sigma_k \models \phi_1 \\ \forall i \leq k < j
\end{multline}

\subsubsection{Mixed Integer Encoding}
In this section, we will detail the mixed integer encoding of Linear Temporal Logic specifications. Let $p$ be an atomic proposition defined over some footstep. Then, this proposition has a corresponding binary variable $P^k$, which is defined to be 1 when $p$ is true at the $k^{th}$ footstep and 0 when $p$ is false at the $k^{th}$ footstep (where $k = 0...N-1$). 

To appropriately extend the semantics of Linear Temporal Logic to mixed integer programming, we redefine the original satisfaction relations outlined in the previous section in terms of integer variables. Let $\sigma_k$ is a finite run that starts at footstep $k$.

Firstly, $ \sigma_k \models p $ can be encoded by the following integer constraint,
\begin{equation}
    P^k = 1
\end{equation}

The conjunction satisfiability relation, $ \sigma_k \models \bigwedge_{j=1}^{m} \phi_j $, where m is any integer greater than 1, can be encoded by the following integer constraint,
\begin{equation}
    \sum_{j=1}^{m} P^k_{\phi_j} = m
    \label{eqn:conj1}
\end{equation}

The negation satisfiability relation, $ \sigma_k \models \lnot \phi $ can be encoded by the following integer constraint,
\begin{equation}
    P^k_{\phi} = 0
    \label{eqn:neg1}
\end{equation}

Although the disjunction satisfiability relation, $ \sigma_k \models \bigvee_{j=1}^{m} \phi_j $, where m is any positive integer greater than 1, can easily be derived from the conjunction (Eq. \ref{eqn:conj1}) and negation (Eq. \ref{eqn:neg1}) satisfiability relations by means of Eq. \ref{eqn:disjunc_equiv}, we propose a more succinct encoding:
\begin{equation}
    \sum_{j=1}^{m} P^k_{\phi_j} >= 1
    \label{eqn:satdisj}
\end{equation}

Finally, $ \sigma_k \models \bigcirc \phi $ can be encoded by the following integer constraint,
\begin{equation}
    P^{k+1}_{\phi} = 1
    \label{eqn:satnext}
\end{equation}

The encoding of the until operator, $U$ can be defined in terms of Eqs. \ref{eqn:sat1}-\ref{eqn:satnext} above. Specifically, the \textit{expansion laws} of LTL formulae \cite{baier2008principlesmodel} give us the following recursive identity for the until operator, $U$:
\begin{equation}
    \phi_1 U \phi_2 = \phi_2 \vee ( \phi_1 \wedge \bigcirc ( \phi_1 U \phi_2 ))
\end{equation}

Granted that footstep planning involves a finite horizon, any loops in the plan are finite and would not lead to circular reasoning (as pointed out in \cite{2006biereltlencodings}). As a result, we can avoid using an auxiliary encoding of $U$, which involves under-approximating its functionality \cite{wolff2014optimization}. Instead, our proposed approach formulates a set of nested mixed-integer constraints (using Eqs. \ref{eqn:sat1}-\ref{eqn:satnext}) over a vector of binary variables, in which each variable corresponds to the satisfiability of $\phi_1 U \phi_2$ at a particular foothold in the run. 

Let $T \in \{0,1\}^{1 \times (N-k)}$ be a vector of binary variables of length $N-k+1$. Assuming Eqs. \ref{eqn:sat1}-\ref{eqn:satnext}, $\sigma_k \models \phi_1 U \phi_2$ can be encoded by the following set of integer constraints: 
\begin{equation}
    T^{i} = P_{\phi_2}^{i} \vee (P_{\phi_1}^{i} \wedge \bigcirc T^{i+1}) \hskip 2em i = k, ..., N-1
    \label{eqn:until_main}
\end{equation}

with base case,

\begin{equation}
    T^{N} = P_{\phi_2}^{N}
    \label{eqn:until_base}
\end{equation}

and satisfiability constraint (i.e. $\sigma_k$ must satisfy the LTL specification, $\phi_1 U \phi_2$)

\begin{equation}
    T^{k} = 1
    \label{eqn:until_sat}
\end{equation}

Instead of deriving the mixed-integer encodings for safety, liveness, and persistence LTL formulae from the general definitions above, we can more efficiently encode them as follows:

\textit{Safety (``always"):} The safety LTL specification, $\sigma_k \models \square \phi$ can be encoded as the following integer constraint:

\begin{equation}
    \sum_{i=k}^{N} P_{\phi}^{i} = N-k+1
    \label{eqn:safety}
\end{equation}

\textit{Liveness (``eventually"):} The liveness LTL specification, $\sigma_k \models \lozenge \phi$ can be encoded as the following integer constraint:

\begin{equation}
    \sum_{i=k}^{N} P_{\phi}^{i} \geq 1
    \label{eqn:liveness1}
\end{equation}

\textit{Liveness (``repeated eventually"):} The liveness LTL specification, $\sigma_k \models \square \lozenge \phi$ (repeated eventually) can be encoded as the following set of integer constraints:

\begin{equation}
    \sum_{i=j}^{N} P_{\phi}^{i} \geq 1 \hskip 2em j = k,...,N
\end{equation}

\textit{Persistence (``eventually always"):} The persistence LTL specification, $\sigma_k \models \lozenge \square \phi$ (eventually always) is defined with the help of Eq. \ref{eqn:satdisj}, and is encoded as follows:

\begin{equation}
    \bigvee_{j=k}^{N} \sum_{i=j}^{N} P_{\phi}^{i} = N-j+1 
\end{equation}

\section{Examples}
Our mixed integer encodings facilitate the integration of LTL specifications in the footstep planner. This is particularly powerful in ensuring safe performance of region-based locomotion tasks. Specifications can detail desired locomotion behavior such as the requirement that the robot must eventually enter a set of regions or that it must reach some region within some number of footsteps. Specifications can also be used to outline underlying environment (region-specific) characteristics such as the requirement that if a footstep is placed in a particular region, the following footsteps must remain in that region or that access to certain regions is only granted provided particular regions have already been visited.

In this section, we present three LTL-constrained footstep planning scenarios. These scenarios were solved using the commercial optimization solver, Gurobi on a laptop with a 2.3 GHz quad-core processor and 16 GB of memory. Source code and animations of these scenarios can be found on Github at \href{https://github.com/vikr53/ltl_footstep_planner}{{\fontfamily{cmr}\selectfont
https://github.com/vikr53/ltl\_footstep\_planner
}}. 

\subsection{Liveness}
We will first consider encoding a region-based liveness specification that necessitates that certain obstacle-free regions be visited eventually. The planner then solves for an optimal set of footsteps that fulfil this requirement while also reaching some goal footstep position and orientation. 

Let $p_{R_3}$ and $p_{R_4}$ be the atomic propositions, ``a footstep is in Region 1" and ``a footstep is in Region 2" respectively.

\textit{LTL Formula: }
\begin{equation}
    \lozenge (p_{R_3} \vee p_{R_4})
\end{equation}

\textit{Mixed-Integer Encoding: }
Using the liveness (Eq. \ref{eqn:liveness1}) and disjunction (Eq. \ref{eqn:satdisj}) encodings, we can express the above LTL specification as the following linear mixed-integer constraint:

\begin{equation}
    \sum^{N}_{j=1} (H_{R_3,j} + H_{R_4,j}) \geq 1
\end{equation}

\begin{figure}
  \includegraphics[width=\linewidth]{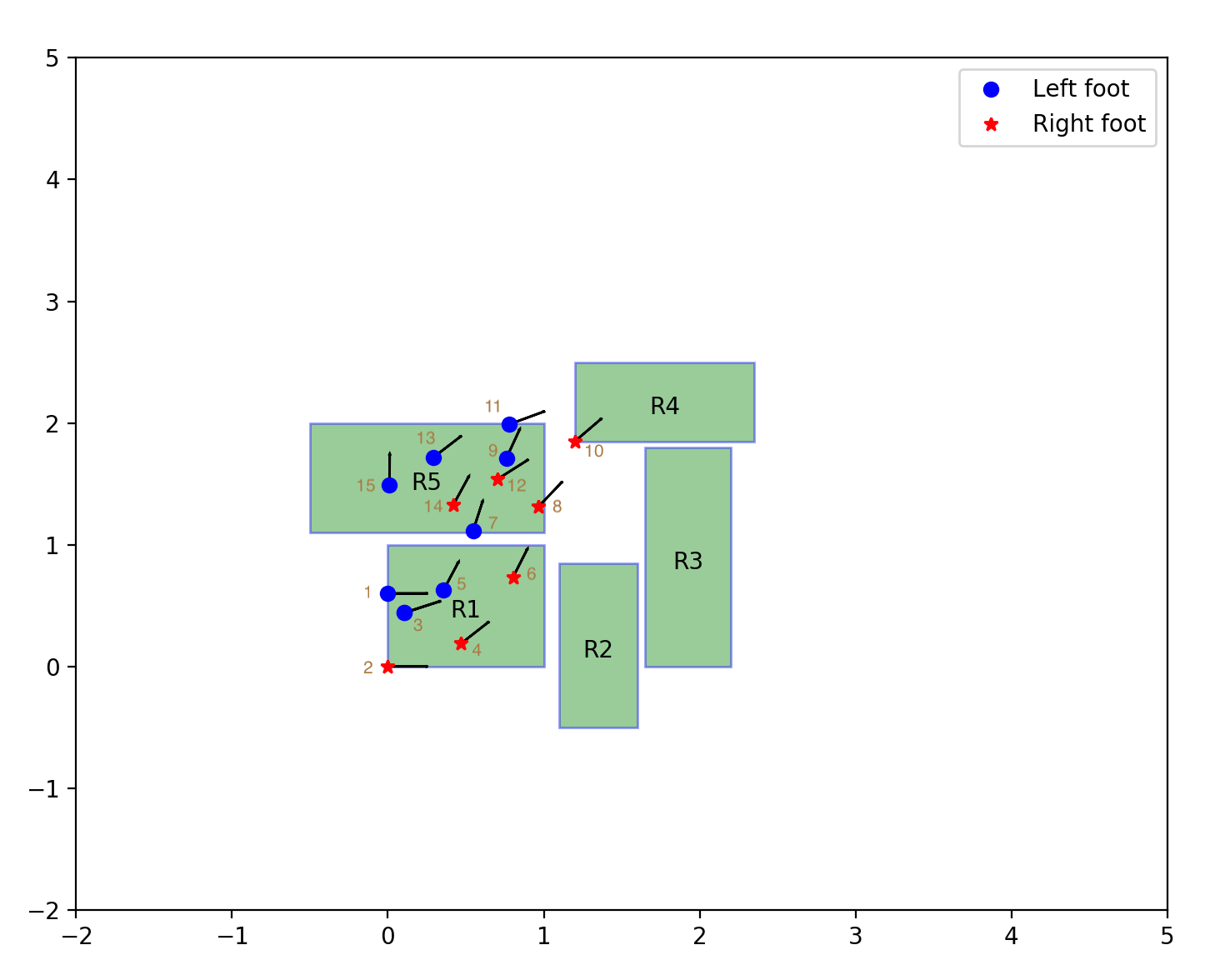}
  \caption{\textit{Scenario 1. } Encoded LTL Specification: $\lozenge (p_{R_3} \vee p_{R_4})$. Right-leg footsteps are shown in red stars and left-leg footsteps are shown in blue circles. The orientation of each footstep is depicted by its protruding black arrow. Each footstep is numbered in brown. Goal footstep location and orientation: [$x$: 0, $y$: 1.5, $\theta$: $\pi/2$]}
  \label{fig:scenario_1}
\end{figure}

The result from the planner is depicted in Figure \ref{fig:scenario_1}, where the red stars are the right leg footsteps, the blue circles are the left leg footsteps and the green boxes are the obstacle-free convex regions. It also must be noted that the first two footholds are assumed to be fixed (i.e. the humanoid robot is in some initial double-support stance). In this scenario, the planner must eventually reach either region 3 (``R3") or region 4 (``R4") before reaching its goal state - ($x$: 0, $y$: 1.5, $\theta$: $\pi/2$). As is evident from the simulation, the planner is able to reason that the most efficient way to satisfy this requirement is to visit region 4 through region 5 before returning to its goal state in region 5. This optimal result is solved for in around 25s. 

\subsection{Until}
We now demonstrate the encoding of the until, $U$, temporal operator in the footstep planner. The until operator can be used to ensure that the approach to some goal state remains within certain pre-defined desirable regions. This has applications in a wide-variety of locomotive tasks such as surveying and reconnaissance. 

Let $p_{R_1}$, $p_{R_2}$, and $p_{R_3}$ be the atomic propositions, ``a footstep is in Region 1", ``a footstep is in Region 2" and ``a footstep is in Region 3" respectively. The length of the plan is $N = 13$.

\textit{LTL Formula: }
\begin{equation}
    (p_{R_1} \vee p_{R_2}) U p_{R_3}
\end{equation}

\textit{Mixed-Integer Encoding: }
First, we define a vector of binary variables, $T \in \{0,1\}^{1\times N}$. Then, we specify the base case (as given by Eq. \ref{eqn:until_base}):
\begin{equation}
    T^{N} = H_{R_3, N}
\end{equation}

where R3 denotes the index for Region 3 in matrix $H$. Next, following the formulation defined in Eq. \ref{eqn:until_main} and Eq. \ref{eqn:satdisj}, we first encode $ \phi_1^k \equiv p_{R_1}^k \vee p_{R_2}^k $, where $k \in {1,...,N-1}$, and $p_{R_1}^k$ and $p_{R_2}^k$ are the atomic propositions, ``the $k^{th}$ footstep is in Region 1", ``the $k^{th}$ footstep is in Region 2".

\begin{equation}
    H_{R_1,k} + H_{R_2,k} \geq 1 - M (1-P_{\phi_1}^k)
\end{equation}
\begin{equation}
    H_{R_1,k} + H_{R_2,k} \leq 1 - m + M (P_{\phi_1}^k)
\end{equation}

where $m$ and $M$ are sufficiently small and large positive integers respectively. Then, using the big-M formulation and the conjuction encoding given by Eq. \ref{eqn:conj1}, we encode the term, $\phi_1 \wedge \bigcirc(\phi_1 U \phi_2)$:

\begin{equation}
    P_{\phi_1}^k + T^{k+1} \geq 2 - M (1 - B^k)
\end{equation}
\begin{equation}
    P_{\phi_1}^k + T^{k+1} \leq 2 - m + M (B^k)
\end{equation}

where $B$ is a vector of binary variables, i.e. $B \in {0,1}^{1\times \{N-1\}}$. Finally, we have the following two constraints to finish encoding the entire specification:

\begin{equation}
    H_{R_3,k} + B^k \geq 1 - M (1-T^k)
\end{equation}
\begin{equation}
    H_{R_3,k} + B^k \leq 1 - m + M (T^k)
\end{equation}

All that remains is simply adding the satisfiability constraint (Eq. \ref{eqn:until_sat}):

\begin{equation}
    T^{1} = 1
\end{equation}

\begin{figure}
  \includegraphics[width=\linewidth]{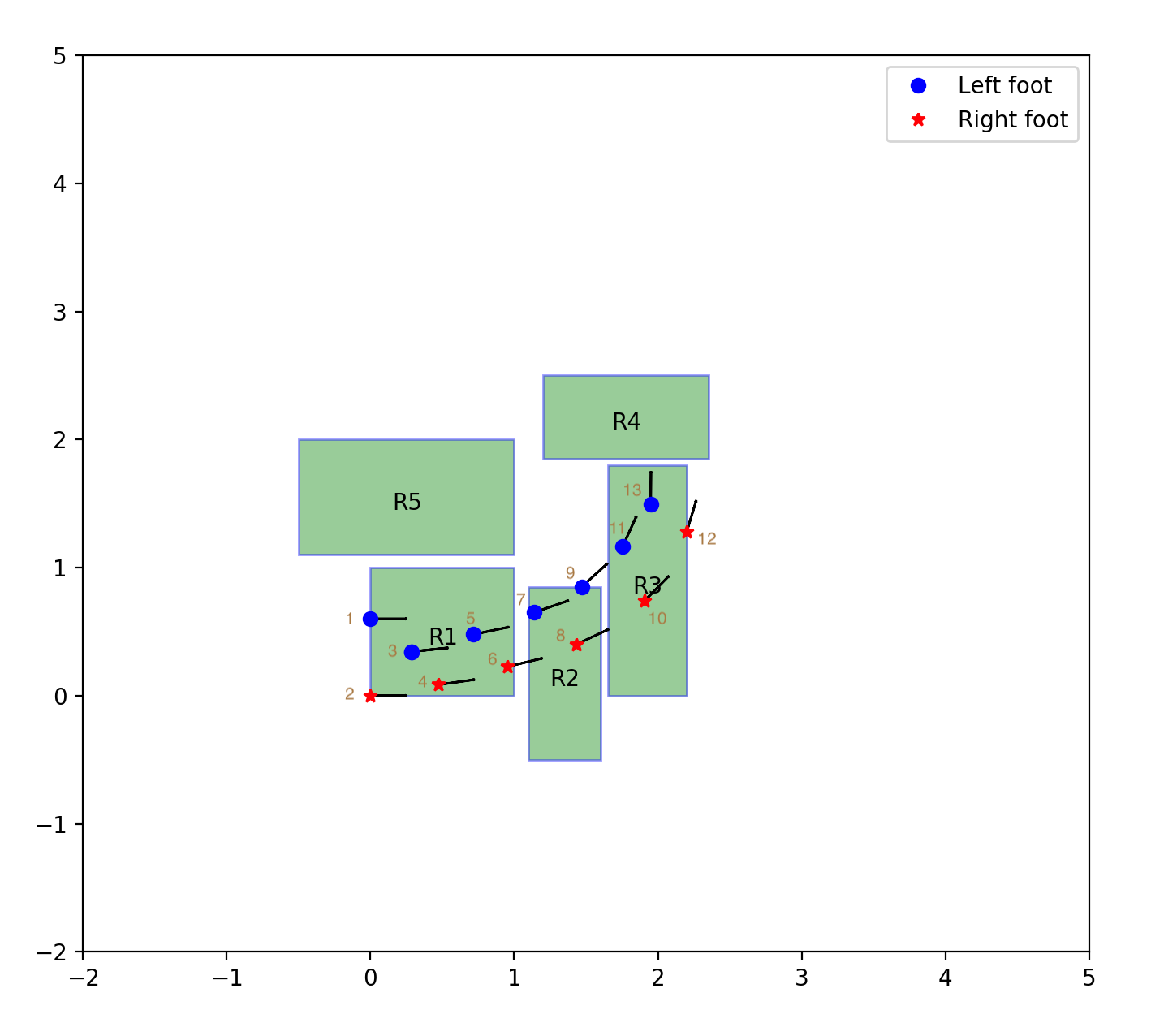}
  \caption{\textit{Scenario 2. } Encoded LTL Specification: $(p_{R_1} \vee p_{R_2}) U p_{R_3}$. Right-leg footsteps are shown in red stars and left-leg footsteps are shown in blue circles. The orientation of each footstep is depicted by its protruding black arrow. Each footstep is numbered in brown. Goal footstep location and orientation: [$x$: 2, $y$: 1.5, $\theta$: $\pi/2$]}
  \label{fig:scenario_2}
\end{figure}

Fig. \ref{fig:scenario_2} displays the solved LTL-constrained footstep plan. It is evident that the 13 planned footsteps abide by the given specification, ``until Region 3 is reached, footsteps must either be in Region 1 or Region 2", while reaching its goal state - ($x$: 2, $y$: 1.5, $\theta$: $\pi/2$). This example was solved in around 2s. 

\subsection{Timed Specifications}
In this example, we consider the encoding of timed LTL constraints, where each footstep is assumed to be a single time unit. Defined over a discrete time domain, these specifications are relatively simple to encode but allow for the expression of interesting constraints like ``always be in a particular region or set of regions for the first $p$ steps" or ``eventually reach a region within the $p^{th}$ and $q^{th}$ footstep", where $p, q$ are arbitrary positive integers. For the sake of this example, we consider the following specification:

\textit{LTL Formula: } 
\begin{equation}
    \square^{7 \leq k \leq 15} p_{R_2}
\end{equation}

\textit{Mixed-Integer Encoding: }
Using Eq. \ref{eqn:safety}, we can write the above specification as the following constraint:

\begin{equation}
    \sum_{j=7}^{15} H_{R2,j} \geq 15-7+1
\end{equation}

\begin{figure}
  \includegraphics[width=\linewidth]{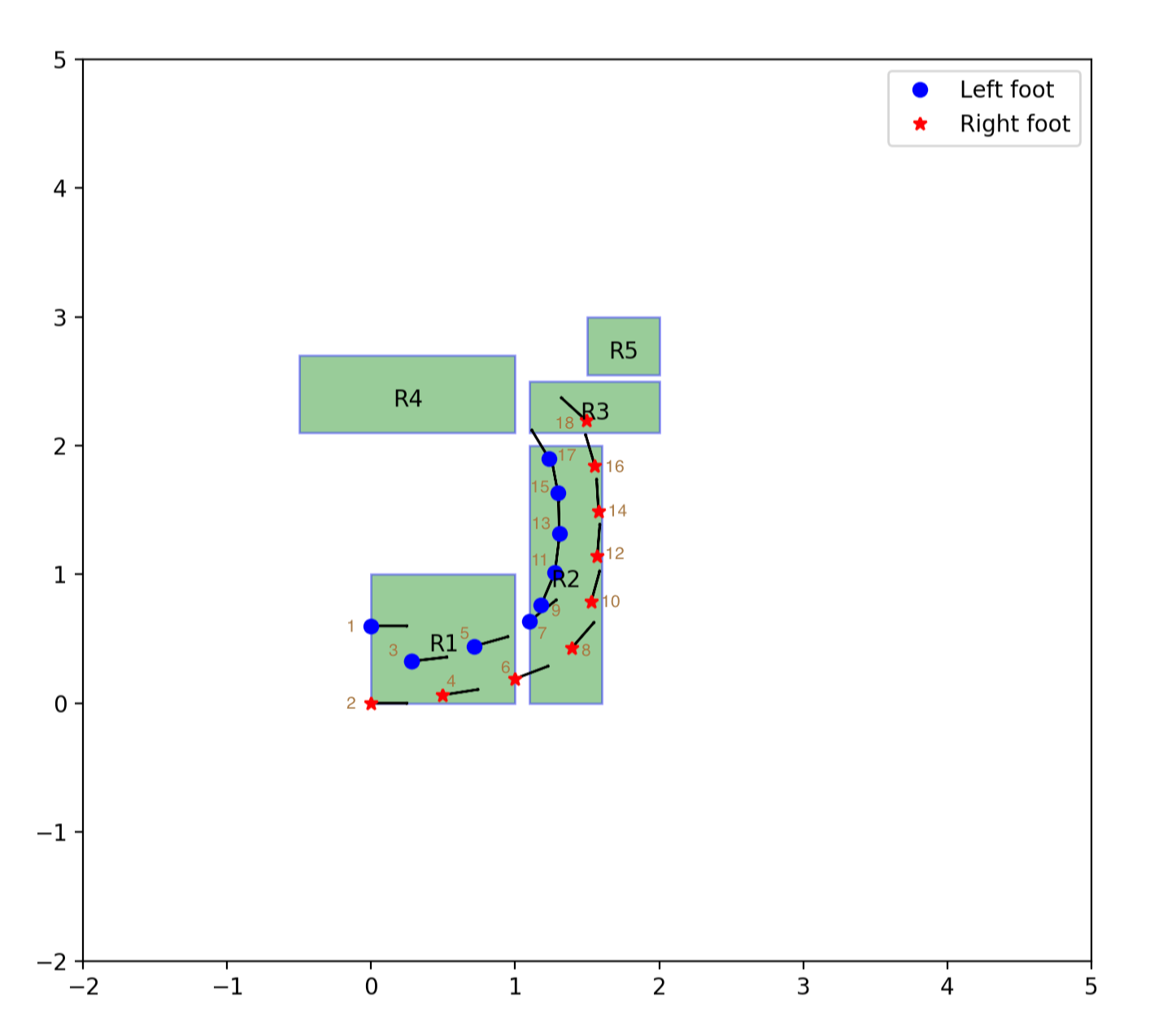}
  \caption{\textit{Scenario 3. } Encoded LTL Specification: $\square^{7 \leq k \leq 15} p_{R_2}$. Right-leg footsteps are shown in red stars and left-leg footsteps are shown in blue circles. The orientation of each footstep is depicted by its protruding black arrow. Each footstep is numbered in brown. Goal footstep location and orientation: [$x$: 1.5, $y$: 2.2, $\theta$: $3\pi/4$]}
  \label{fig:scenario_3}
\end{figure}

The footstep plan generated for this scenario is shown in Fig.  \ref{fig:scenario_3}. The plan's length is $N = 18$ steps. It is evident that the plan satisfies the discrete-timed specification, ``the $7^{th}$ to the $15^{th}$ footstep must be in Region 2", while finally reaching its goal state - ($x$: 1.5, $y$: 2.2, $\theta$: $3\pi/4$). The solve time for this example was around 2s. 

Aside from encoding surveying behaviors, an interesting application of such timed specifications is assessing the accessibility of a region. While taking into account the particular formulation of the footstep planner (the planning horizon, reachability constraints, the obstacle-free regions and other encoded specifications), this specification can determine whether a specific region or set of regions can be reached within some number of steps. For example, in order to determine whether Region 2 can be reached within 5 footsteps for the formulation used for this example, we can simply encode the specification, $\lozenge^{\leq 5} p_{R_2}$, into the optimization program. By nature of the encoding, the planner attempts to satisfy this specification. Thus, if the MIQCQP is rendered infeasible, we know definitively that it is not possible to reach Regon 2 in 5 footsteps, i.e. Region 2 is not accessible in 5 steps with the current formulation. The opposite is true when the MIQCQP is solvable, i.e. Region 2 is accessible in 5 steps.

\section{Ongoing/Future Work}

\subsection{Stride Adjustment}
We have also considered the encoding of region-based stride adjustments. Since the traversibility of terrain can vary significantly over a footstep plan, the planner may need to tread certain regions more carefully than others. Often, the first step in reducing the risk of loosing dynamic stability on account of the terrain is reducing the stride length. This can be easily implemented by changing the reachability constraints for regions with difficult terrain:

From Eq. \ref{eqn:footstep_reach}, $p_i$ and $r_i$, $i \in \{1,2\}$, are the centers and radii of the circles whose intersection defines the footstep reachability of the robot. To define the reduced reachability constraint, we merely have to intersect two relatively smaller and less offset circles with centers and radii, $p_i^s$ and $r_i^s$ $i \in \{1,2\}$ respectively. Then, we use the big-M formulation to encode the implication that smaller strides only occur in specific regions. For example, smaller strides in Region 2 can be encoded as:

\begin{multline}
     \norm{ \begin{bmatrix}x_j \\ y_j \end{bmatrix} - \bigg( \begin{bmatrix}x_{j-1} \\ y_{j-1} \end{bmatrix} + \begin{bmatrix} cos(\theta_j) & -sin(\theta_j) \\ sin(\theta_j) & cos(\theta_j) \end{bmatrix} p_i^s \bigg) } - M (1-H_{R_2,j}) \\ \leq r_i^s
\end{multline}

\begin{figure}
  \includegraphics[width=\linewidth]{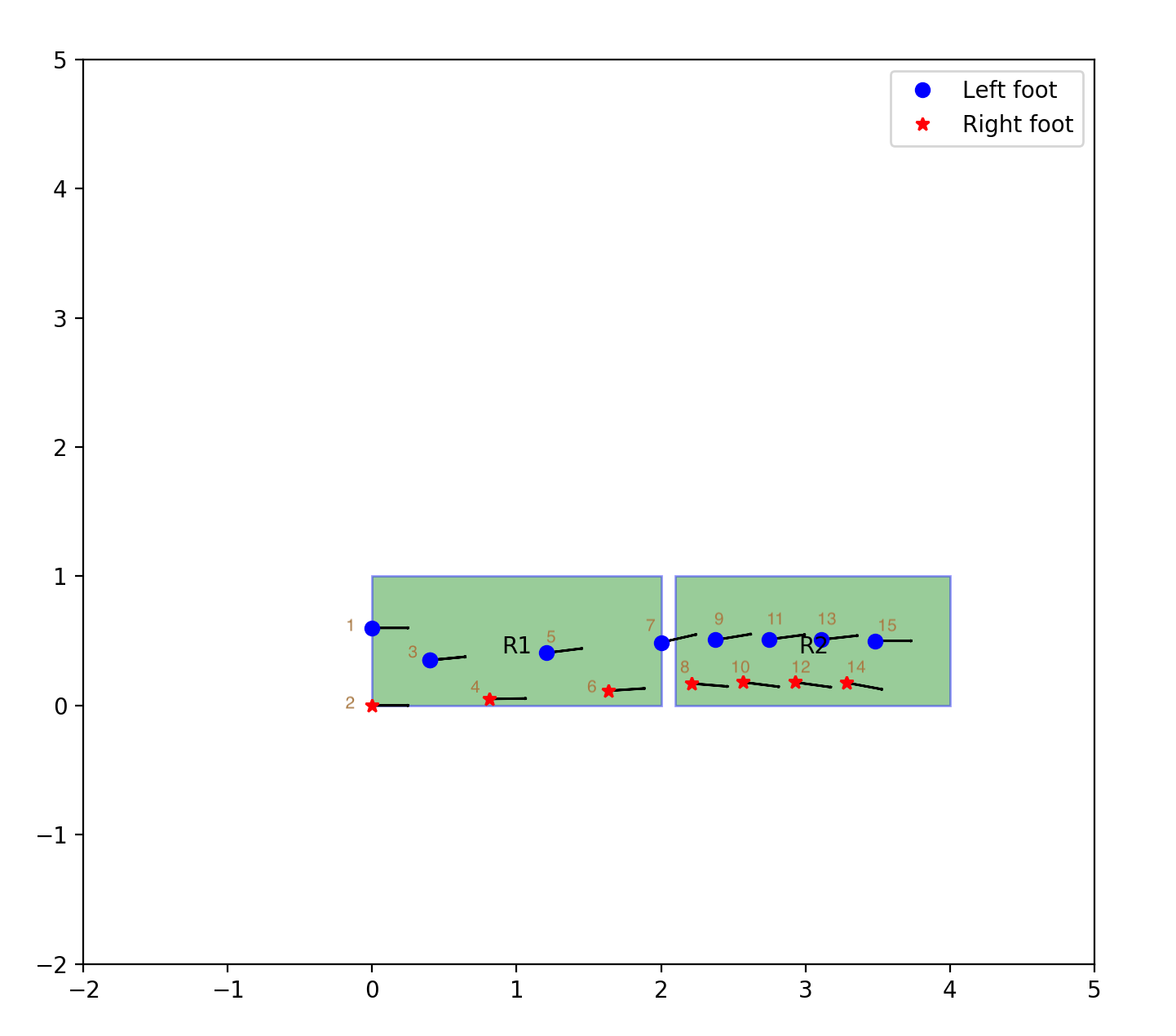}
  \caption{\textit{Stride Adjustment Example.} Planner makes shorter strides in Region 2 since it is assumed to have difficult terrain. Goal footstep location and orientation: [$x$: 3.5, $y$: 0.5, $\theta$: 0]}
  \label{fig:stride_adj}
\end{figure}

Fig. \ref{fig:stride_adj} depicts the result of encoding the above specification in the footstep planner. It is evident that the planner is able to shorten its stride when the robot enters Region 2. As mentioned previously, this has several applications, especially in dealing with difficult terrain. For example, given that a certain region is known to have more slippery terrain, the robot can shorten its stride in this region to reduce the risk of slipping. However, it must be noted that solely planning shorter strides does not guarantee safer locomotion. Appropriate contact forces must also be planned. We hope to investigate the integration of our LTL-constrained planner with trajectory optimization (for planning through contact) \cite{2014posatrajopt}.

\subsection{Multi-Contact Planning}
In addition to specifying region-based locomotion behavior, encoded LTL specifications can potentially be extended to the more difficult multi-contact problem, allowing us to encode higher-level reasoning about the safety and performance of these plans. We propose that this is possible by first defining the planner in three dimensions, then approximating the COM position as the center of the support polygon and adding approximate arm reachability constraints. Lastly, LTL constraints can be defined over the ordering of contacts. Given the combinatorial complexity of the multi-contact problem, this formulation would be able to not only make the problem more tractable but also guarantee the safety and performance of multi-contact plans. We are actively researching this idea and have done some preliminary tests on the simpler footstep planner (which has only 2 end-effectors).

\textit{Preliminary Results: }
To demonstrate that the ordering of end-effectors in contact need not be defined explicitly (as was done in the formulation of the original MIQCQP footstep planner \cite{deits2014footstep}), we encode ordering-related LTL specifications into the planner and let the planner formulate the appropriate end-effector contact sequence. In the case of the footstep planner, the safe and desirable ordering of end-effectors in contact is trivial - the right foot must follow the left foot, and a footstep can be made by either the left or right foot, not both. This is encoded as the linear mixed-integer constraints:

\begin{equation}
    LL^j + RL^j = 1
\end{equation}

\begin{equation}
   -M (1-LL^{j-1}) + RL^j \leq 1
\end{equation}

\begin{equation}
   M (1-LL^{j-1}) + RL^j \geq 1
\end{equation}

where $LL^j$ and $RL^j$, for $j \in \{2,...,N\}$, are binary variables that determine whether the left leg and right leg make the $j^{th}$ footstep respectively. In addition, the reachability constraint in Eq. \ref{eqn:footstep_reach} must be changed to adjust to whether a particular footstep was chosen to be made with the left or right foot:

\begin{multline}
     \norm{ \begin{bmatrix}x_j \\ y_j \end{bmatrix} - \bigg( \begin{bmatrix}x_{j-1} \\ y_{j-1} \end{bmatrix} + \begin{bmatrix} cos(\theta_j) & -sin(\theta_j) \\ sin(\theta_j) & cos(\theta_j) \end{bmatrix} p_1 \bigg) } - M (1-LL^j) \\ \leq r_1
\end{multline}

\begin{multline}
     \norm{ \begin{bmatrix}x_j \\ y_j \end{bmatrix} - \bigg( \begin{bmatrix}x_{j-1} \\ y_{j-1} \end{bmatrix} + \begin{bmatrix} cos(\theta_j) & -sin(\theta_j) \\ sin(\theta_j) & cos(\theta_j) \end{bmatrix} p_2 \bigg) } - M (1-RL^j) \\ \leq r_2
\end{multline}

This encoding was successful and generated similar results to the original formulation. It must be noted that while these specifications are in effect equivalent to original implicit encoding for the case of the footstep planner, it is neither preferable nor feasible to add ordering constraints on a contact-by-contact basis for the more complicated multi-contact problem. Instead, encoding such LTL specifications allow the planner to not only determine the optimal sequence of contacts but also guarantee that this sequence follows some safety constraints. In the multi-contact planning problem, these constraints could include simple requirements like $\square (p_{lleg} \vee p_{rleg})$ (one foot should always be on the ground), and even criteria for the necessity of multi-contact plans (such as uncertainty in the stability of certain contacts). This idea seems promising, and we hope to make progress on its development soon. The code for this example is available as the script, {\fontfamily{cmr}\selectfont
footstep\_planner\_contact\_ordering\_specs.py
} in the Github repository linked earlier in this paper. 

\subsection{Exploiting Sparse Constraint Matrices}
While this planner can solve our footstep planning problem for 10 steps, 5 regions and a few LTL constraints in less than a second, increasing the planning horizon (i.e. planning for more footsteps), introducing new regions and encoding additional constraints increase the complexity of solving the mixed-integer program. Hence, a limitation of this planer is its poor scalability.

We propose that the sparsity of the $H$, $S$ and $C$ matrices can possibly be exploited to increase the solve times of plans with large horizons (greater than 20 footsteps). However, this is merely an idea at the moment, and we hope to explore this in more detail soon.


\bibliographystyle{IEEEtran}
\bibliography{IEEEabrv,IEEEexample}

\end{document}